\documentclass[letterpaper]{article}
\usepackage{natbib,alifeconf}
\usepackage{amsmath}
\usepackage{subfig}
\usepackage{hyperref}
\usepackage{appendix}
\usepackage{balance}

\title{Energy Costs and Neural Complexity Evolution in Changing Environments}
\author{Sian Heesom-Green$^{1}$, Jonathan Shock$^{1,2,3}$ \and Geoff Nitschke$^{1}$ \\
\mbox{}\\
$^1$University of Cape Town, South Africa \\
$^2$INRS Montreal. Canada \\
$^3$NiTheCS, Stellenbosch\\ South Africa\\
hsmsia001@myuct.ac.za, jonathan.shock@uct.ac.za, gnitschke@cs.uct.ac.za}

\begin{document}
\maketitle

\begin{abstract}
The \textit{Cognitive Buffer Hypothesis} (CBH) posits that larger brains evolved to enhance survival in changing conditions. However, larger brains also carry higher energy demands, imposing additional metabolic burdens. Alongside brain size, brain organization plays a key role in cognitive ability and, with suitable architectures, may help mitigate energy challenges. This study evolves \textit{Artificial Neural Networks} (ANNs) used by Reinforcement Learning (RL) agents to investigate how environmental variability and energy costs influence the evolution of neural complexity, defined in terms of ANN size and structure. Results indicate that under energy constraints, increasing seasonality led to smaller ANNs. This challenges CBH and supports the \textit{Expensive Brain Hypothesis} (EBH), as highly seasonal environments reduced net energy intake and thereby constrained brain size. ANN structural complexity primarily emerged as a byproduct of size, where energy costs promoted the evolution of more efficient networks. These results highlight the role of energy constraints in shaping neural complexity, offering \textit{in silico} support for biological theory and energy-efficient robotic design.
\end{abstract}

\section{Introduction}

The evolution of the brain is a fascinating topic that has been widely investigated, yet much remains to be understood. Many studies have investigated why some animals have evolved larger brains than others \citep{sayol2016environmental, sol2009revisiting, michaud2022impact}. The \textit{Cognitive Buffer Hypothesis} (CBH) suggests that large brains evolved to improve adaptability and enhance survival in changing conditions, such as seasonal environments \citep{allman1993brain, sol2009revisiting, michaud2022impact}. However, larger brains are typically metabolically costly, and it is not always feasible for organisms to increase their energy intake \citep{sayol2016environmental, michaud2022impact, smaers2013brain}. The \textit{Expensive Brain Hypothesis} (EBH) highlights this constraint, proposing that an increase in brain size must be met by an increase in net energy intake or reduced energy allocation to other vital organs \citep{isler2009expensive}. Despite these limitations, organisms often face the need to adapt to environmental changes \citep{smaers2013brain}, raising the question of how changing environments and energy costs impact neural complexity evolution.

Alongside brain size, the organization of the brain plays a crucial role in cognitive ability \citep{cohen2016segregation}. Brain structures that balance functional segregation, where information is processed within specialized neural groups, and integration, where these groups communicate, are key biomarkers for diverse cognitive function \citep{tononi1994measure,cohen2016segregation}. While larger brains demand more energy, which organisms may struggle to obtain, an optimally structured brain could achieve similar cognitive functions with lower energy costs \citep{smaers2013brain}. This structural efficiency may be crucial in mitigating the energetic constraints associated with larger brains \citep{smaers2013brain, oizumi2014phenomenology}. This raises the question: \textit{Does neural evolution favor larger brains or more efficient wiring under energy constraints?} Hereafter, we use the term neural complexity to refer to the combined influence of brain size and structural organization.

In neuro-evolution, the artificial evolution of \textit{Artificial Neural Networks} (ANNs), provides a valuable tool for studying neural complexity evolution \citep{miikkulainen2025neuroevolution}. While both brain size and structural organization contribute to neural complexity, they are often studied independently. For example, previous studies in evolutionary robotics have defined neural complexity in terms of ANN size and investigated whether imposing complexity costs, such as energy penalties, evolves more efficient robot controllers \citep{nagar2019cost, hallauer2020energy}, with less focus on how structural organization in neural controllers might adapt to balance these constraints. Similarly, studies on neural complexity in terms of network structure typically investigate conditions driving evolution towards more complex structures, but with constrained ANN size conditions \citep{edlund2011integrated, joshi2013minimal, yaeger2006evolution,cg2018effect}. 

Thus this study investigates neural complexity evolution for ANN size and structure, given testing the CBH remains little investigated using neuro-evolution as an experimental tool. Also, while CBH focuses on how changing environments influence the evolution of brain size, little research has examined its impact on evolving neural structures. 

\paragraph{Research Objective:} Specifically, this study investigates how changing environments and energy costs impact the evolution of neural complexity, defined by both ANN size and structure. To do this, we evolve ANNs of Reinforcement Learning (RL) agents in four environments, ranging from static to highly seasonal (1 to 4 seasons), under two energy regimes (with and without energy costs that scale with ANN size). 
We thus test predictions of the CBH to examine the relative importance of ANN size versus structural complexity in adaptation, including whether energy costs drive the evolution of more efficient controllers. 

\section{Methods}

This section details the task, agent neuro-evolution, energy costs, task performance and neural complexity metrics.

\subsection{Task Environment}
\label{sub:env}
\begin{figure}[t]
    \centering
    \includegraphics[scale=0.31]{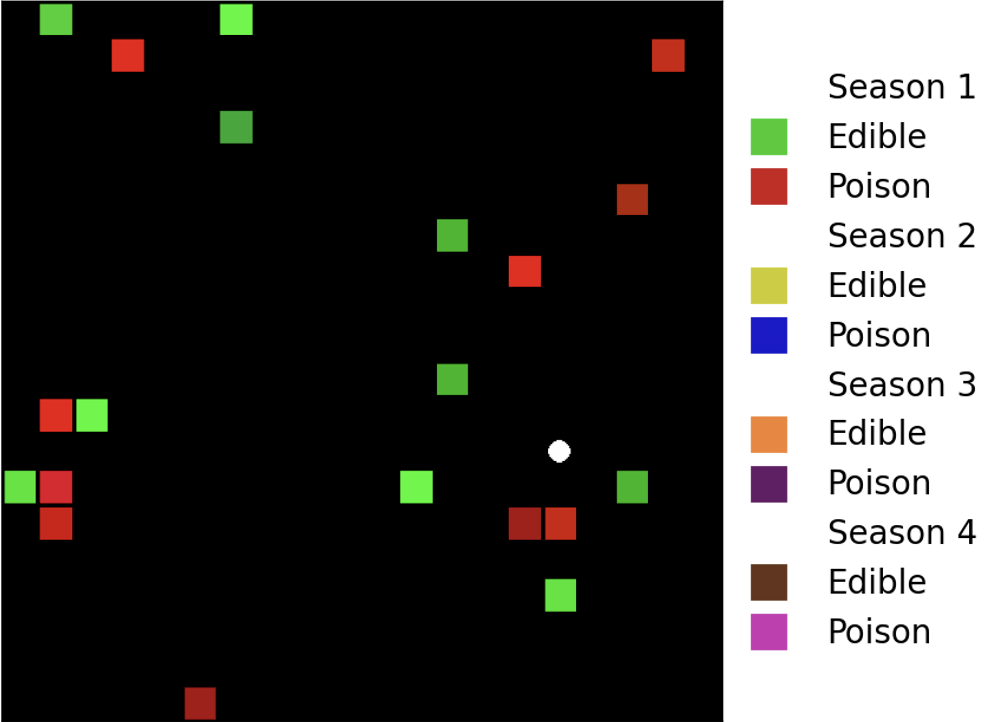}
    \caption{20×20 grid-world. Cell color indicates food type (edible or poisonous). The agent is the white circle. The legend shows seasonal food-color mappings.
    }
    \label{fig:env}
\end{figure}

Agents operate in a 20×20 2D environment (Figure~\ref{fig:env}), consuming edible foods and avoiding poisonous ones, with edibility signaled by color. Agents expend energy each time step, gain energy from edible foods, and lose energy from poisonous ones (Table~\ref{tab:evolparams}). At the start of each episode, 10 edible and 10 poisonous food items are randomly placed in the environment, along with the agent’s location. When a resource is consumed, it is replaced with a new item of the same type at a random location. Agents have five discrete actions (move up, down, left, right, or eat) and observe a 9×9 window of nearby cell colors, encoded as 243 RGB values (9×9×3), normalized to [0, 1]. Each episode ends after 100 time steps (Table~\ref{tab:evolparams}).

\paragraph{Agent Rewards:} The reward process mirrors the energy dynamics that agents experience over their lifetime.
\begin{itemize}
\item -0.01 at each time step (reflects energy expenditure)\footnote{-0.01 is the baseline energy expenditure per step when not scaled by ANN size (see Equation~\ref{eq:EC}).}.
\item +1 upon consuming edible food (energy replenishment).
\item -1 upon consuming poisonous food (energy depletion).
\end{itemize}

\paragraph{Food Colors and Seasonal Changes:}
Environments differ in the number of seasons per episode (1–4), with each season defined by a unique mapping of colors to food edibility (see Figure~\ref{fig:env} legend). At the start of each experimental run (i.e., per simulation seed), a predefined set of colors (green, yellow, orange, brown, red, blue, purple, pink) is shuffled: the first four are assigned as edible (for seasons 1–4), and the rest as poisonous. This mapping remains fixed across all episodes and generations within a run but varies between simulation seeds. 

Each 100-step episode is evenly divided across seasons, with agents experiencing the same fixed seasonal sequence. For example, a 2-season environment consists of 50 steps of season 1 followed by 50 of season 2. Agents receive no explicit cues about season transitions and must infer them through experience, but these changes follow a fixed sequence in every episode of a run and are therefore predictable rather than stochastic. Additionally, each food item's final color is generated by sampling each RGB channel within a 0.2 range of its base color.

\subsection{Lifetime Learning via Reinforcement Learning}\label{par:RL}

Agents undergo lifetime learning per generation using RL, where during their lifetime, they adapt by adjusting ANN controller connection weights given environment interactions. This lifetime learning mirrors natural adaptive behavior, potentially enhancing agent ability to adapt to environmental changes \citep{doncieux2015evolutionary, urzelai2001evolution}. RL is guided by the \textit{Proximal Policy Optimization} (PPO) algorithm \citep{schulman2017proximal}. 
Each RL agent has an actor and a critic network that share the same ANN topology, which evolves via NEAT \citep{stanley2002evolving}, but differ in their output layers. The actor network learns the optimal policy, while the critic network estimates the value function.

\begin{itemize}
    \item \textbf{Input layer} (243 nodes): Encodes the current state as a feature vector, representing the RGB color values of each cell in the agent’s 9×9 field of view.
    
    \item \textbf{Critic output layer} (1 node): State-value function \( V(s) \), estimating the expected cumulative reward from state \( s \).
    
    \item \textbf{Actor output layer} (5 nodes): Represents the policy \( \pi(a|s) \), defining the probability distribution over the agent’s five possible actions in state \( s \).
\end{itemize}

The RL hyper-parameters\footnote{RL hyper-parameter details \href{https://github.com/sianmay/ECNCECE/blob/main/neat_config_exp_v2}{here}.} and the actor’s ANN evolve via NEAT. The critic network is derived by replacing the actor’s output layer with a single state-value node $V(s)$, which inherits all incoming output node connections.

\begin{table}[t]
\centering \caption{Simulation and evolution parameters}
\begin{tabular}{|l|c|}
\hline
\textbf{Experiment Parameters}      &                 \\ \hline
Number of runs                      & 20              \\
Episode length (time-steps)         & 100             \\ 
Agent energy usage (per time-step)  & -E             \\ 
Energy gain (edible food)           & +1             \\ 
Energy loss (poisonous food)        & -1             \\ 
\hline
\textbf{Simulation Parameters}      &                 \\ \hline
Grid-world size                     & 20x20           \\
Observation / action space size     & 243 / 5         \\ \hline
\textbf{Neuro-evolution Parameters} &                 \\ \hline
Population size                     & 150             \\
Generations                         & 400             \\
ANN architecture                    & Feedforward     \\
ANN initial connectivity            & Fully connected \\
Hidden nodes                        & 1               \\ \hline
\textbf{Mutation Parameters}        &                 \\ \hline
Weight mutation rate                & 0.1             \\
Node add / delete probability       & 0.5 / 0.0       \\
Connection add / delete probability & 0.5 / 0.5       \\ \hline
\end{tabular}
\label{tab:evolparams}
\end{table}

\subsection{Agent Neuro-evolution}\label{subsec:neuroevol}

The neuro-evolution process uses the NEAT algorithm to evolve ANN topologies and weights \citep{stanley2002evolving}, outlined in the following:

\begin{enumerate}
\item \textbf{Initialization}: 150 RL agents (individuals) are generated with random RL hyper-parameters and a fully connected (initial structural complexity $N_C=0$; Equation \ref{eq:Nc}) feedforward ANN with one hidden node. This setup reduces initial network size (fewer connections than no hidden layer). Weights and biases are initialized from $\mathcal{N}(0,1)$.

\item \textbf{Lifetime Learning}: Before evaluation, RL agents update their inherited ANN weights over 1000 episodes (100 000 time steps), using their inherited hyper-parameters and the PPO algorithm \citep{schulman2017proximal}. Importantly, any updates to ANN weights during this learning phase are not inherited by the next generation (Baldwinian Evolution).

\item \textbf{Evaluation}: After the learning phase, assess each individual's fitness based on RL performance (total reward per episode), averaged over 100 episodes (using seeds not seen during learning).

\item \textbf{Selection, Reproduction, and Replacement}: The top 10\% (15 individuals) are selected based on fitness to produce offspring via crossover and mutation, modifying inherited architectures, initial weights (pre-learning), and RL hyper-parameters. The fittest two are preserved. The remaining 148 are replaced by offspring.
\end{enumerate}

Steps 2-4 are repeated for 400 generations, with each generation using the same fixed training and evaluation environment seed split within a given simulation run. Further details on the evolutionary parameters\footnote{Full NEAT evolutionary parameter details \href{https://github.com/sianmay/ECNCECE/blob/main/neat_config_exp_v2}{here}.} used in this neuro-evolution process are provided in Table \ref{tab:evolparams}. Evolutionary parameters were optimized via Bayesian hyper-parameter optimization, and connection and node addition and deletion rates were manually tuned over values \{0.1-0.5\} to balance task performance and complexity variance. Node deletion rate was set to 0, since disconnected nodes could still be removed.

\subsection{Task Performance Metric: Net Energy Intake}\label{sub:taskperf}

The task performance metric differs from the fitness function (accumulated rewards over an episode) because it excludes energy expenditure (E), and primarily focuses on net energy intake, measured as the number of edible foods consumed minus the number of poisonous foods consumed during the agent's lifetime. Excluding energy costs from the task performance calculation enables a better comparison between agents with different energy expenditures, that is, those with energy costs on ANN size versus those without. This still influences the neuro-evolution process and enables us to investigate the impact of higher energy costs for larger ANNs on the evolution of neural complexity.

\paragraph{Rationale for Energy Budget Design}

Agents were not assigned an initial energy budget because energy dynamics did not affect their lifespan or trial duration. This setup was intentionally chosen to avoid biases when comparing performance between the different energy expenditure conditions. In particular, it avoids the issue noted by \cite{hallauer2020energy}, where energy depletion shortened agent lifespans, giving agents without energy costs more time to perform the task. Since agents in this study could act for the full duration regardless of energy level, assigning an initial energy value was unnecessary.

\renewcommand{\arraystretch}{1.3}
\begin{table*}[hbt!]
\caption{Task Environments (a) and Experiment Sets (b) \textit{(NEC = No Energy Costs; EC = Energy Cost)}}\label{tab:exp_sets}
\centering
\begin{tabular}{c c}
(a) Task Environments & (b) Experiment Sets\\ 
\begin{tabular}{|l|l|}
\hline
\textbf{Environment} & \textbf{Seasons} \\ \hline
1 (Static)                       & 1                \\ \hline
2                        & 1,2              \\ \hline
3                        & 1,2,3            \\ \hline
4                        & 1,2,3,4          \\ \hline
\end{tabular}
&
\begin{tabular}{|c|c|c|c|}
\hline
\textbf{Experiment Set} &
  \textbf{Energy Expenditure (E)} &
  \textbf{\begin{tabular}[c]{@{}c@{}}Experimental \\ Variables\end{tabular}} &
  \textbf{\begin{tabular}[c]{@{}c@{}}Evaluation \\ Metrics\end{tabular}} \\ \hline
1 (NEC) &
  -0.01 &
  Environments: 1-4 &
  \begin{tabular}[c]{@{}c@{}}Neural Complexity\\ Task Performance\end{tabular} \\ \hline
2 (EC) &
  $-0.01 \times \dfrac{N_S^{\text{current}}}{N_S^{\text{gen0}}}$ &
  Environments: 1-4 &
  \begin{tabular}[c]{@{}c@{}}Neural Complexity\\ Task Performance\end{tabular} \\ \hline
\end{tabular}

\end{tabular}
\end{table*}
\renewcommand{\arraystretch}{1}

\subsection{Neural Complexity Metrics}\label{sub:nc}

\paragraph{ANN size ($\mathbf{N_S}$):} Larger ANNs are often associated with the potential for more complex behaviors due to their increased number of free parameters \citep{lehman2011abandoning}. Therefore, ANN size is commonly defined as the total number of free parameters (connections and non-input nodes) in the network \citep{nagar2019cost, hallauer2020energy, lehman2011abandoning, nitschke2017evolutionary, yu2010network}, and is calculated using Equation \ref{eq:Ns}:

\begin{equation} \label{eq:Ns}
    N_S = \text{(\# connections)} + \text{(\# non-input nodes)}
\end{equation}

\paragraph{ANN structural complexity ($\mathbf{N_C}$):}
In neuroscience, two core principles of brain functional organization are \textit{segregation} (specialized processing within groups of neurons) and \textit{integration} (efficient information exchange between these groups) \citep{sporns2013network,cohen2016segregation}.
A balance between segregation and integration has been shown to support diverse cognitive abilities and rich, flexible dynamics \citep{tononi1994measure,deco2015rethinking}. Neural complexity ($N_C$) is considered high when this balance is achieved, and low when networks are either fully segregated or fully integrated \citep{tononi1994measure}. \citet{tononi1994measure}’s original $N_C$ measure, a precursor to Integrated Information Theory (IIT) measures, becomes computationally infeasible for networks larger than ~20 nodes, as it requires evaluating every possible bipartition \citep{toker2019information,mediano2018measuring}.
To address this, we define the $N_C$ metric for evolved ANN controllers as the ratio between modularity (segregation) and global efficiency (integration), both well-established, scalable graph-theoretic measures for functional segregation and integration \citep{toker2019information, palma2025balance, van2014brain, cohen2016segregation, capouskova2022integration}.

\begin{description}
    \item \textbf{Modularity (M):} Measures segregation (range: [0, 1]) \citep{cohen2016segregation}, by comparing within-module connection density to between-module connections \citep{newman2010networks}.  Higher values indicated  greater segregation. Modules were identified using the Louvain community detection algorithm \citep{blondel2008fast}.
    
    \item \textbf{Global Efficiency (E):} Measures integration (range: [0, 1]) \citep{capouskova2022integration}, computed as the average inverse shortest path length between all pairs of nodes \citep{latora2001efficient}. Higher values indicate greater integration through more efficient global communication across the ANN.
\end{description}  

The neural complexity ratio is computed with Equation \ref{eq:Nc}:
\begin{equation}  \label{eq:Nc}
    N_C = \frac{\min{(M(G), E(G))}}{\max{(M(G), E(G))}} 
\end{equation}

Where G is the graph representation of the ANN, analyzed using \textit{NetworkX} \citep{hagberg2008exploring}. This ratio is sensitive to changes in either component, decreasing when one measure dominates the other. When $N_C \approx 1$, segregation and integration are well-balanced (high neural complexity). When $N_C \approx 0$, one property dominates, indicating low neural complexity.

\subsection{Energy Costs}

We consider two energy expenditure scenarios: one without ANN-size-dependent energy costs (NEC) and one where energy costs scale with ANN size (EC).

\begin{description}
        \item \textbf{NEC (No Energy Costs):} Agents have a constant energy expenditure per time step (Equation \ref{eq:NEC}):
    \begin{equation}\label{eq:NEC}
        E=-0.01
    \end{equation}
    Over a 100 time-step trial, total energy loss is $-1$, meaning edible food consumption $(+1)$ replenishes energy, while poisonous food consumption $(-1)$ depletes it. NEC serves as a baseline to evaluate agent performance without additional energy constraints.
    
    \item \textbf{EC (Energy Costs):} Energy expenditure scales with ANN size (Equation \ref{eq:EC}):
    \begin{equation}\label{eq:EC}
        E=-0.01\times C
    \end{equation}
    Where, $C$ is the ratio of the current ANN size ($N_S$) to its initial size at generation 0 ($N_S=254$), and where there is a selective pressure imposed against larger networks.
\end{description}

\section{Experiments}

Experiments\footnote{\url{https://github.com/sianmay/ECNCECE}} investigate the impact of energy costs and changing environments on neural complexity evolution.

\paragraph{The Environment Set:} Consists of four increasingly dynamic environments that differ in the number of seasons an agent experiences over its lifetime (Table \ref{tab:exp_sets}).

\paragraph{Experiment sets:} Two experiment sets (Table \ref{tab:exp_sets}) reflecting different energy scenarios, NEC (No Energy Costs, Equation \ref{eq:NEC}) and EC (Energy Costs, Equation \ref{eq:EC}), are used to investigate the impact of changing environments and energy costs on neural complexity evolution, where both evaluate neural complexity and task performance metrics of the fittest agent evolved per environment, averaged over 20 runs.

Comparisons \emph{within} each experiment set examine how changing environments influence neural complexity evolution. This shows if environmental variation alone drives changes in network size and structure, or only when coupled with energy constraints. Comparisons \emph{between} the two experiment sets evaluate how energy costs on ANN size impact neural complexity and task performance versus no energy costs, highlighting trade-offs and their overall effect on agent task performance across environments.

\begin{figure*}[hbt!]%
    \centering
    \subfloat[\centering ANN size ($N_S$, Equation \ref{eq:Ns})]{{\includegraphics[scale=0.50]{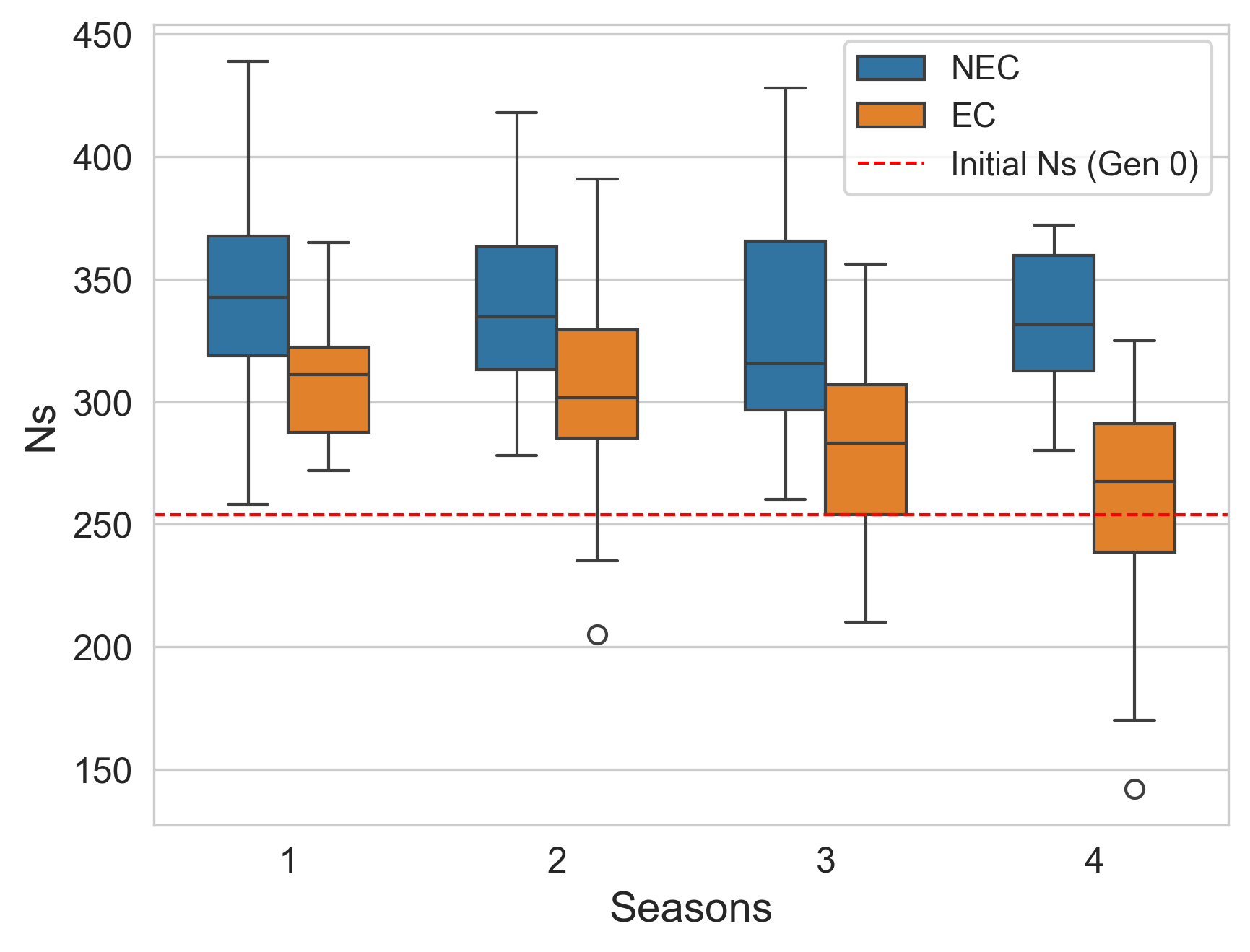} }}%
    \qquad
    \subfloat[\centering ANN size ($N_S$) over Generations]{{\includegraphics[scale=0.50]{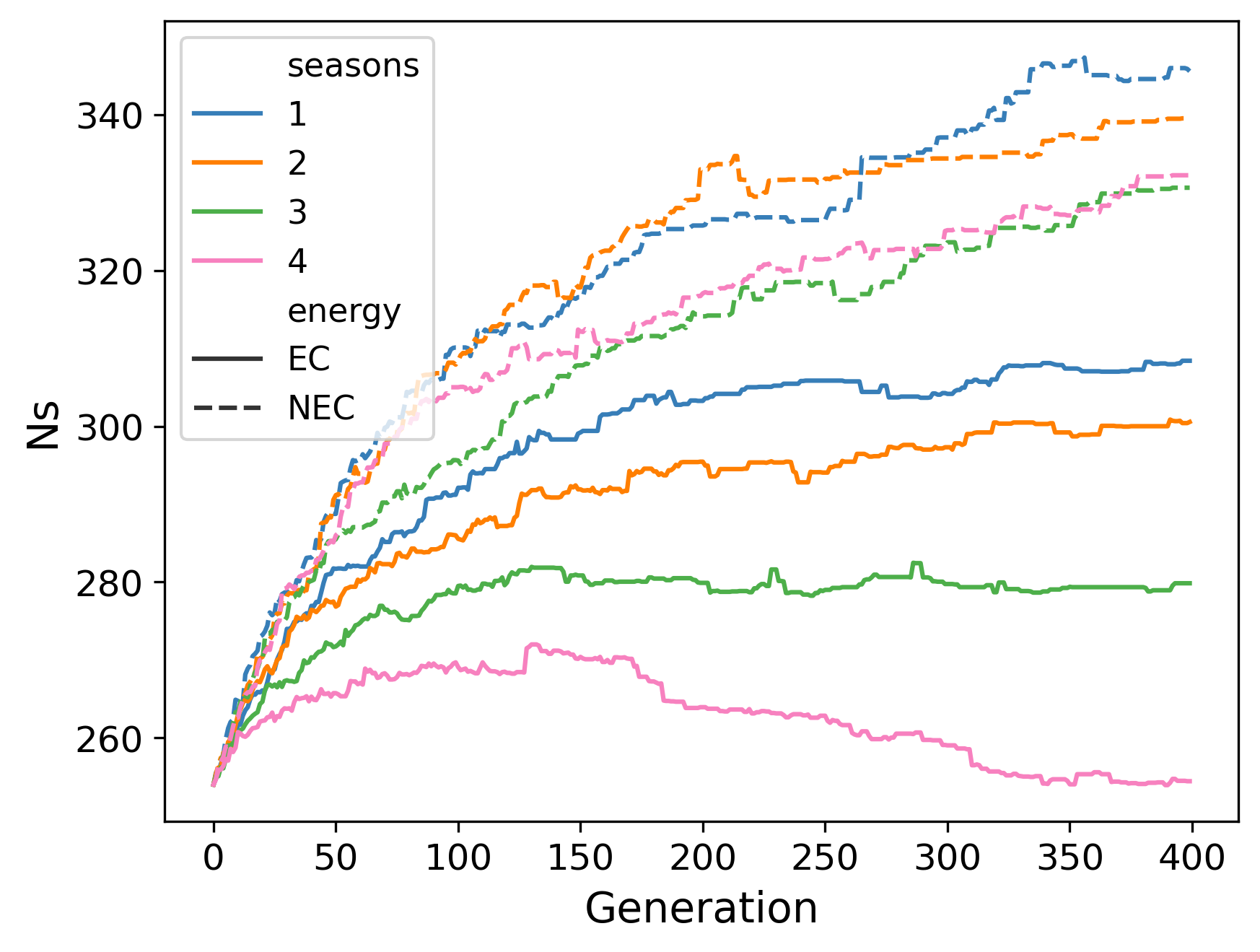} }}%
    \qquad
    \vspace{0.1cm}
    \subfloat[\centering ANN structural complexity ($N_C$, Equation \ref{eq:Nc})]{{\includegraphics[scale=0.50]{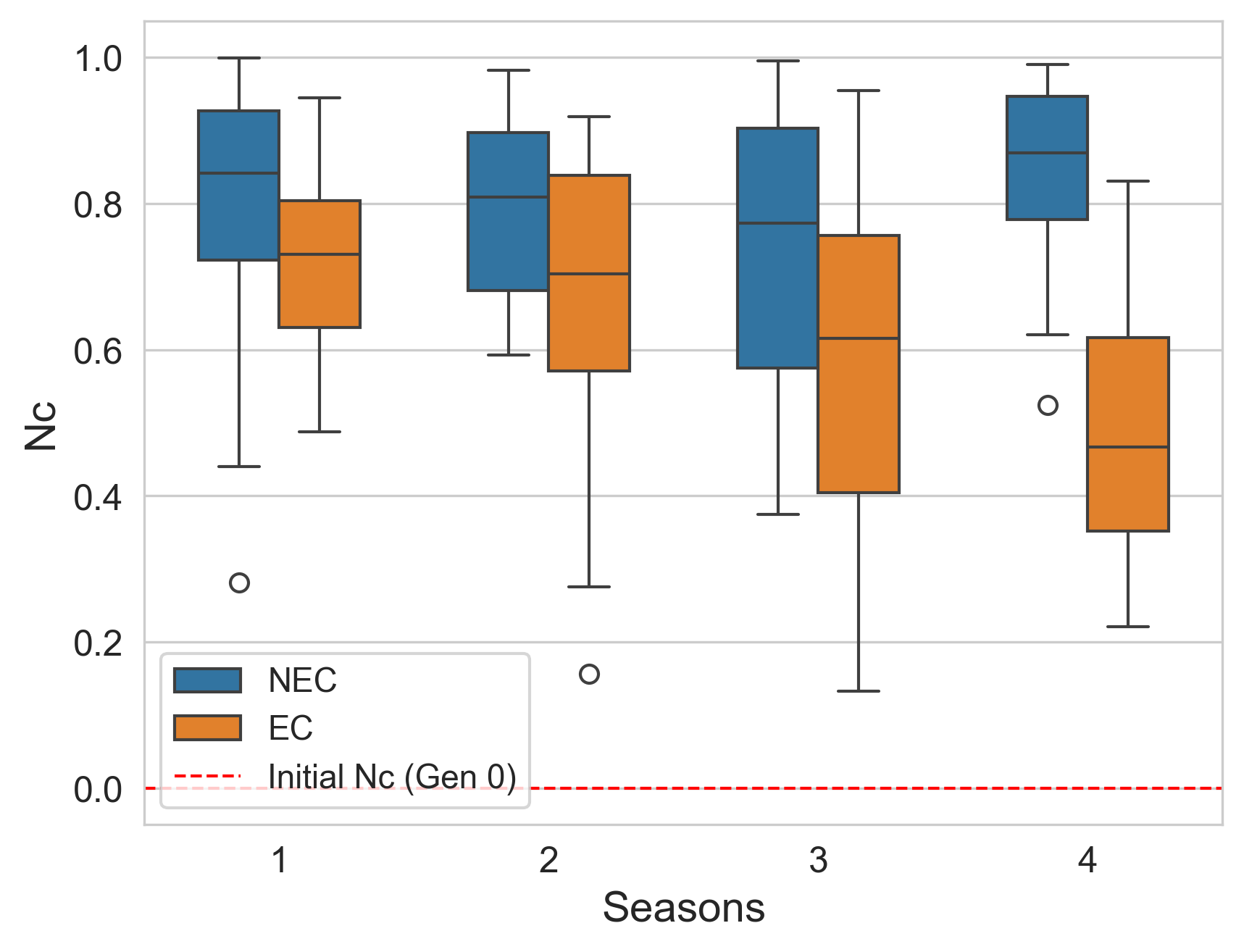} }}%
    \qquad
    \subfloat[\centering ANN structural complexity ($N_C$) over Generations]{{\includegraphics[scale=0.50]{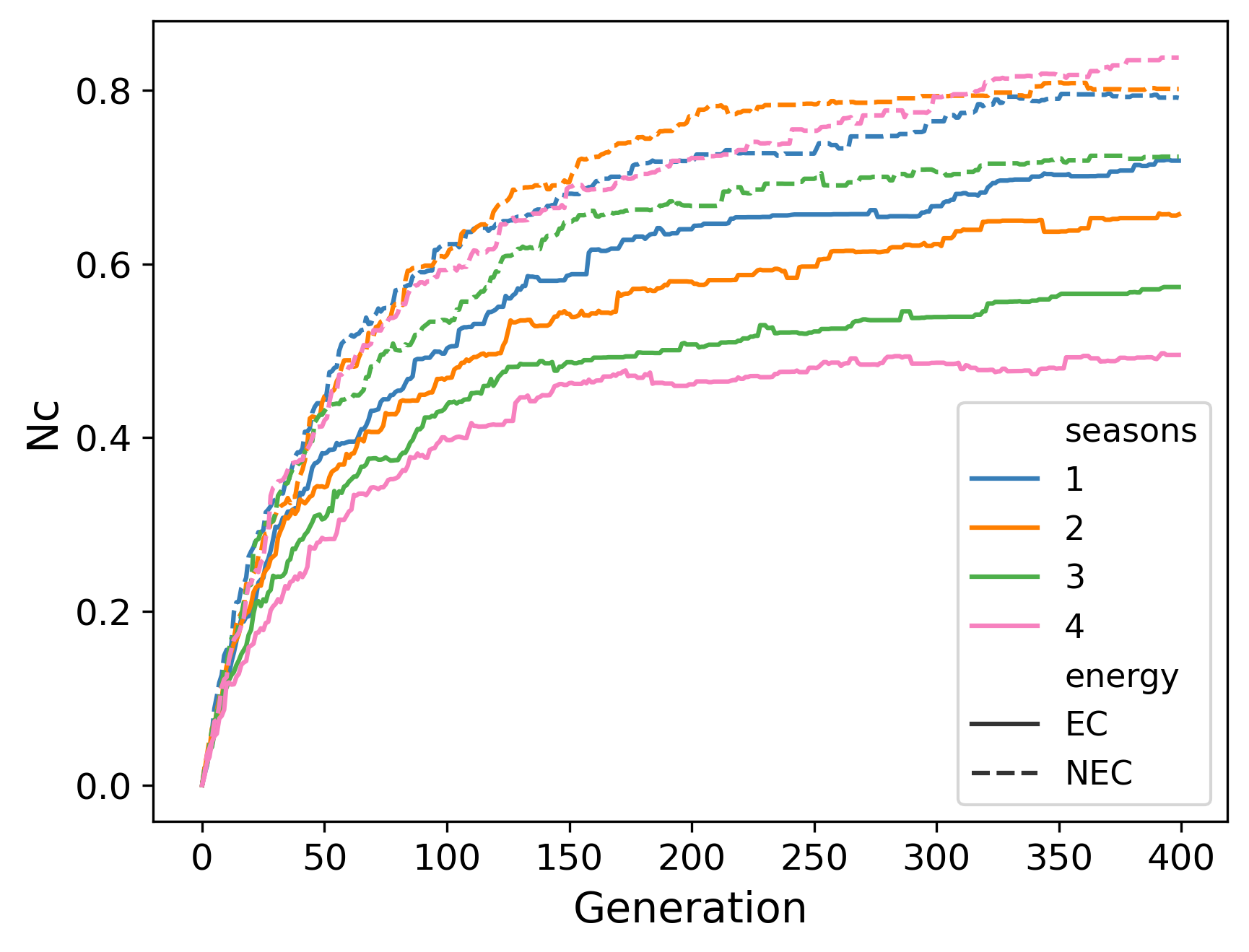} }}%
    \qquad
    \subfloat[\centering Task performance (net energy intake)]{{\includegraphics[scale=0.50]{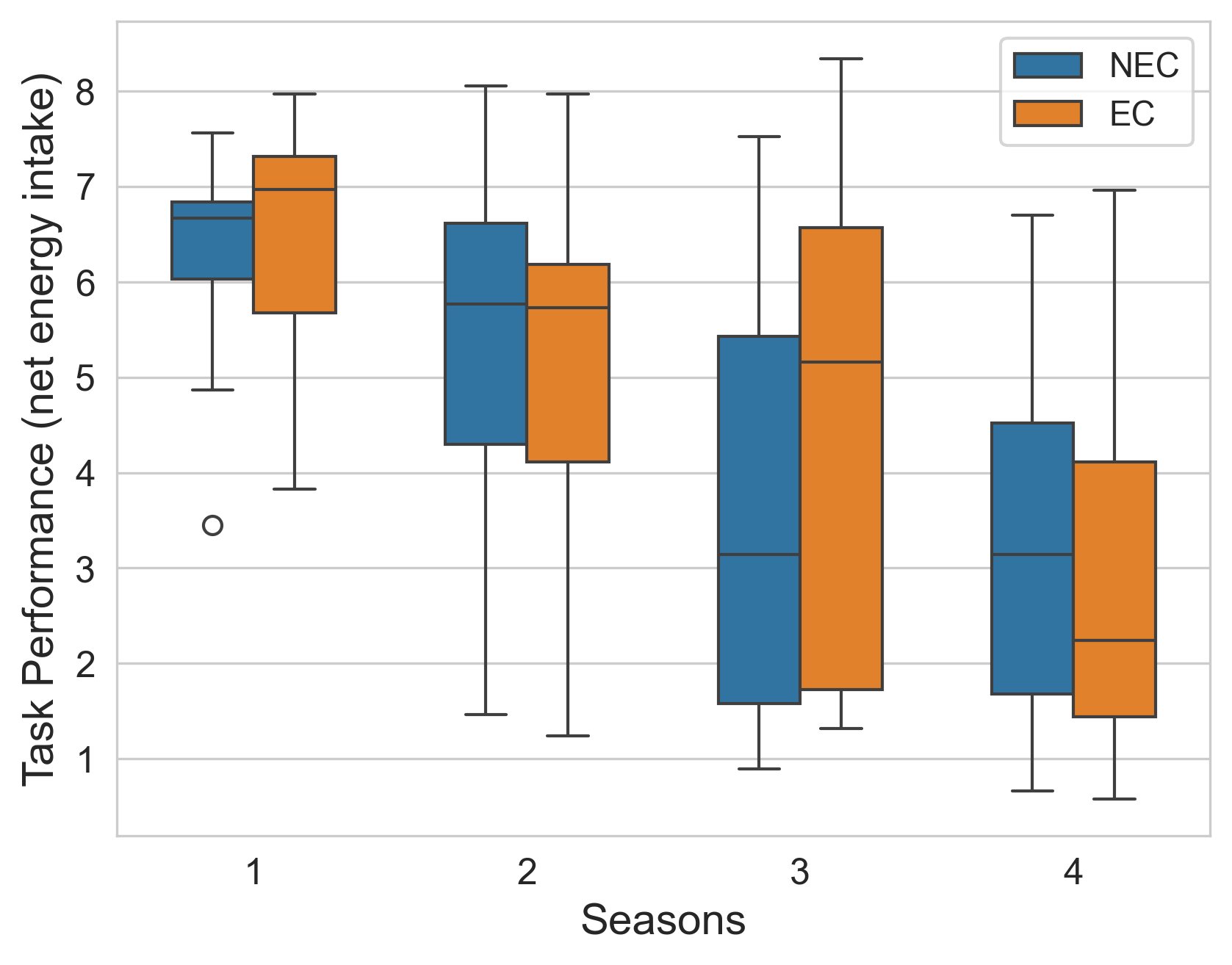} }}%
    \qquad
    \subfloat[\centering Task performance (net energy intake) over Generations]{{\includegraphics[scale=0.50]{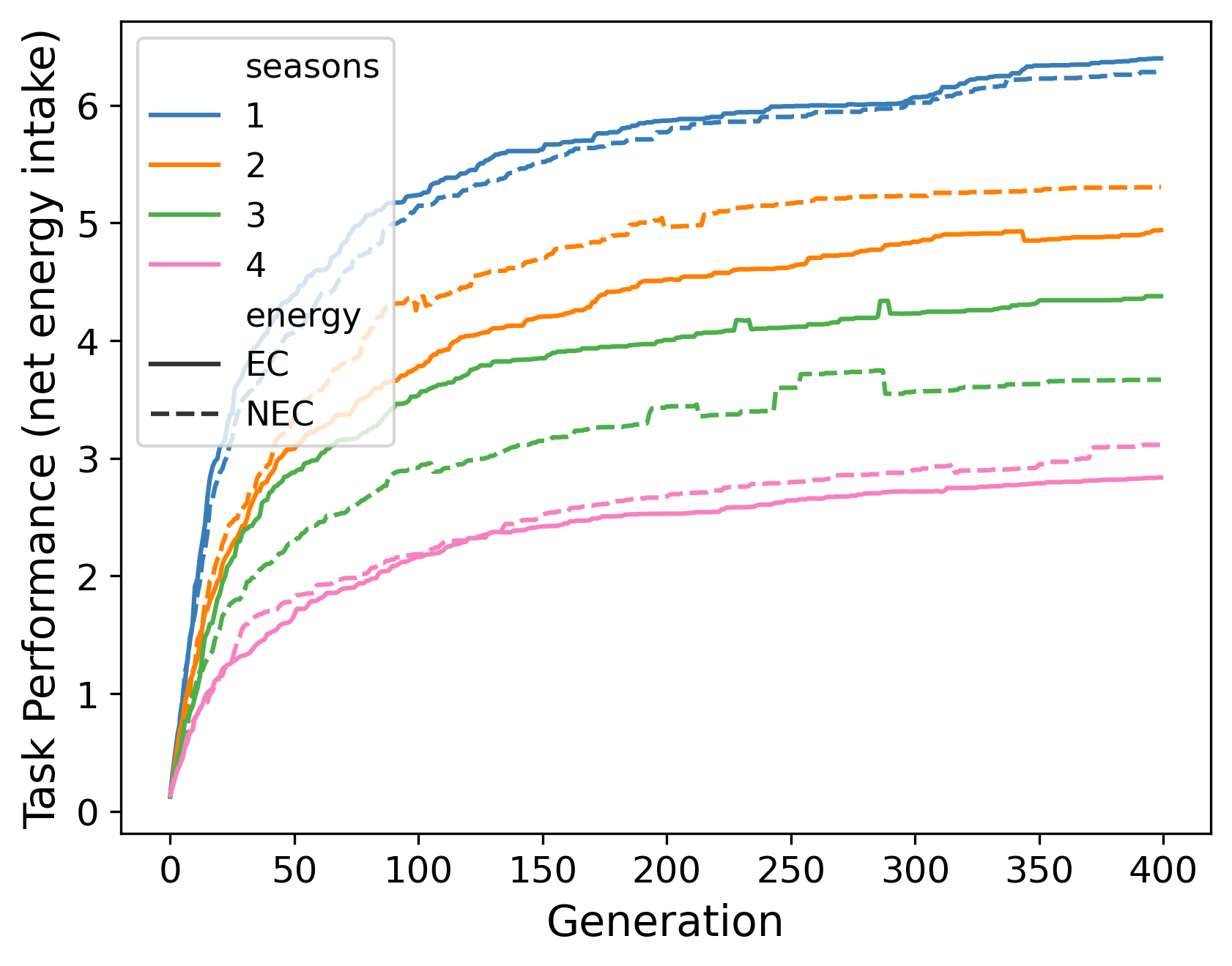} }}
    \caption{\textbf{Left:} Box plots for neural complexity metrics and task performance. The x-axis represents the environments, labeled by the number of seasons, from the static 1-season environment to the most dynamic 4-season environment. The y-axis displays neural complexity metrics / task performance for the fittest evolved ANNs from the final generation (fitness evaluated after lifetime learning using the same evaluation seeds as during evolution), averaged over 20 runs. \textbf{Right:} Neural complexity metrics and task performance over evolutionary time of the current fittest genome, averaged over 20 runs. 
\textit{(NEC = No Energy Costs; EC = Energy Costs)}
}%
    \label{fig:boxplots}%
\end{figure*}

\begin{table*}[t]
\centering
\caption{Experiment set 2 (EC): Post-hoc Dunn test results (Bonferroni-corrected) for pairwise comparisons of ANN size ($N_S$) and structural complexity ($N_C$) across environments with increasing seasonal changes. Significant p-values $(p < 0.05)$ in bold.}
\begin{tabular}{|c|ccc|ccc|}
\hline
\textbf{}       & \multicolumn{3}{c|}{\textbf{$\mathbf{N_S}$}}        & \multicolumn{3}{c|}{\textbf{$\mathbf{N_C}$}}        \\ \hline
\textbf{Seasons} & 2 seasons & 3 seasons & 4 seasons       & 2 seasons & 3 seasons & 4 seasons       \\ \hline
1 season        & 1.0000    & 0.1875    & \textbf{0.0028} & 1.0000    & 0.3051    & \textbf{0.0040} \\
2 seasons       &           & 0.6017    & \textbf{0.0167} &           & 1.0000    & \textbf{0.0468}          \\
3 seasons       &           &           & 1.0000          &           &           & 0.8836          \\ \hline
\end{tabular}
\label{tab:Dunn_EC}
\end{table*}

\begin{table*}[t]
\centering
\caption{Mediation analysis testing whether task performance mediates the relationship between environmental variability (number of seasons) and ANN size ($N_S$). The indirect effect (ab) is significant, with a 95\% confidence interval (CI) of [-151.91, -60.63], indicating that task performance mediates the relationship between environmental variability and $N_S$.}
\begin{tabular}{|l|l|l|l|}
\hline
\textbf{\begin{tabular}[c]{@{}l@{}}a: Effect of n\_seasons on Task \\ Performance (2, 3, 4 seasons)\end{tabular}} &
  \textbf{\begin{tabular}[c]{@{}l@{}}b: Effect of Task \\ Performance on $\mathbf{N_S}$\end{tabular}} &
  \textbf{ab: Indirect Effect} &
  \textbf{95\% CI for ab} \\ \hline
-1.4585 (2), -2.0205 (3), -3.5610 (4) &
  14.5020 &
  -102.0939 &
  {[}-151.91, -60.63{]} \\ \hline
\end{tabular}
\label{tab:mediation}
\end{table*}

\begin{table}[htb!]
\centering
\caption{Spearman rank correlation coefficients reflecting the strength and direction of monotonic trends between generation number and neural complexity metrics ($N_S$ and $N_C$) for each environment and energy scenario. Bold values indicate statistically significant correlations ($p$ $<$ 0.05).}
\begin{tabular}{|c|cc|cc|}
\hline
\textbf{} & \multicolumn{2}{c|}{\textbf{No Energy Cost (NEC)}} & \multicolumn{2}{c|}{\textbf{Energy Cost (EC)}} \\
Seasons & Gen-$N_S$          & Gen-$N_C$          & Gen-$N_S$          & Gen-$N_C$          \\ \hline
1       & \textbf{0.5551} & \textbf{0.6127} & \textbf{0.4627} & \textbf{0.6603} \\
2       & \textbf{0.5414} & \textbf{0.6454} & \textbf{0.3288} & \textbf{0.5319} \\
3       & \textbf{0.4531} & \textbf{0.5343} & \textbf{0.1308} & \textbf{0.4675} \\
4       & \textbf{0.6036} & \textbf{0.7521} & 0.0034          & \textbf{0.4449} \\ \hline
\end{tabular}
\label{tab:spearman_EC}
\end{table}

\section{Results \& Discussion}

Figure \ref{fig:boxplots} presents neural complexity and task performance results (averaged over 20 runs) for the fittest individuals evolved across environments with increasing seasonality, under both energy scenarios (NEC, EC). 
To evaluate differences in evolved complexity across environments, Kruskal–Wallis \citep{kruskal1952use} tests ($p$ $<$ 0.05) were conducted and Dunn’s post-hoc tests \citep{dunn1964multiple} with Bonferroni correction for pairwise comparisons (Table \ref{tab:Dunn_EC}). Spearman rank correlations \citep{spearman1961proof} were used to examine associations between seasonality and neural complexity, as well as evolutionary trends over generations (Table \ref{tab:spearman_EC}). Finally, Mann–Whitney U tests \citep{mann1947test} compared evolved neural complexity and task performance between the NEC and EC scenarios (Table \ref{tab:Energy_Impact}).

\subsection{No Energy Cost (NEC) Scenario}

When we investigated whether changing environments alone impact neural complexity evolution, without considering ANN-size-dependent energy costs (NEC scenario), we found no significant differences in evolved ANN size ($N_S$) or structural complexity ($N_C$) across environments.
That is, from the static 1-season environment to the 4-season environment (Kruskal-Wallis, $p$ $>$ 0.05). Across all conditions, ANNs increased in both $N_S$ and $N_C$ over evolutionary time (Figures \ref{fig:boxplots} (b, d), Table \ref{tab:spearman_EC}). The absence of energy constraints likely allowed unrestricted ANN growth, leading to larger and more complex ANNs regardless of environmental dynamics. These results suggest that environmental dynamics alone do not directly drive the evolution of neural complexity, but instead interact with other selective pressures, such as energy constraints. This result is supported by experimental analysis in evolutionary
biology \citep{fristoe2017big}, elucidating that evolutionary transitions in brain size (for example, in an avian global phylogeny case study), resulted in larger brains evolving with equal likelihood in stable and changing environments. This suggests that other 
environmental and evolutionary factors similarly impact the shaping of neural complexity in both simulation and nature.

\subsection{Energy Cost (EC) Scenario}

The EC scenario further explores the interplay between changing environments, energy constraints and neural complexity evolution by imposing energy costs on larger ANNs. 

\paragraph{Impact of Changing Environments on ANN size ($\mathbf{N_S}$):} When higher energy costs were introduced for larger ANNs, significant differences in evolved ANN sizes ($N_S$) across environments were observed (Kruskal-Wallis, $p$ $<$ 0.05, $\eta^2=0.194$). Specifically, a notable decrease in $N_S$ was found in the 4-season environment compared to both the 1-season and 2-season environments (Table \ref{tab:Dunn_EC}). Spearman correlation tests (Table \ref{tab:spearman_EC}) revealed that $N_S$ modestly increased over generations for the environments with 1 to 3 seasons, but this trend disappeared in the 4-season environment. Figure \ref{fig:boxplots}b, indicates that unlike other environments, the mean $N_S$ of the 4-season environment initially grew but then decreased after approximately 150 generations.

We also confirmed a significant negative association between the number of seasons and $N_S$ ($\rho$ = -0.4322, $p$ $<$ 0.05, Spearman rank correlation coefficient), indicating that increased environmental change is associated with the evolution of smaller networks, refuting the CBH. One explanation is the impact of increasing seasonal changes on task performance ($\rho$ = -0.5389, $p$ $<$ 0.05, Spearman rank correlation coefficient), suggesting that increased environmental dynamics with concomitantly increased task difficulty, leads to reduced consumption of edible food and consequently lower net energy intake. In the current experiment setup, where agents had no competing energetic costs, the EBH implies that sustaining larger ANNs requires higher net energy intake (higher task performance). Thus, environments 
enabling higher agent task performance, such as static environments, are more likely to support the evolution of larger ANNs, which is in line with the EBH.

Furthermore, a bootstrap mediation analysis \citep{preacher2004spss} was conducted to test whether task performance (reflecting energy intake) mediates the relationship between environmental variability (number of seasons) and $N_S$. The analysis indicated a significant negative indirect effect of seasonal variability on $N_S$ (Table \ref{tab:mediation}). Specifically, as the number of seasons increased, task performance decreased (e.g., -1.46 for 2 seasons, -2.02 for 3 seasons, and -3.56 for 4 seasons), and higher task performance was associated with larger $N_S$ (coefficient of 14.50). The indirect effect of -102.09 (95\% confidence interval of [-151.91, -60.63]) indicates that decreasing task performance due to environmental variability leads to smaller $N_S$. This supports the EBH given that environmental change limits ANN size indirectly by reducing performance, thereby limiting energy intake and the feasibility of evolving larger ANNs.

These results align with empirical biological systems research supporting the EBH \citep{isler2009expensive}, including studies on amphibians \citep{luo2017seasonality}, eutherians \citep{van2010effects, graber2017social, van2012large} and marsupials \citep{weisbecker2015evolution}, which similarly show that increased environmental seasonality is often associated with smaller brain sizes due to energy limitations. 

However, ascertaining the exact correlation between ANN size and environmental change, in support of the CBH, remains the topic of ongoing research. 

\paragraph{Impact of Changing Environments on ANN structural complexity ($\mathbf{N_C}$):}
A significant difference was found in evolved ANN structural complexity ($N_C$) across environments (Kruskal-Wallis, $p$ $<$ 0.05, $\eta^2=0.168$), with the 4-season environment resulting in lower $N_C$ than both the 1-season and 2-season environments (Table \ref{tab:Dunn_EC}). We also confirmed a significant negative association between the number of seasons and $N_C$ ($\rho$ = -0.4062, $p$ $<$ 0.05, Spearman rank correlation), indicating that increased environmental change is also generally associated with the evolution of less complexly structured networks. However, while ANN size did not significantly increase over generations in the 4-season environment
(Figures \ref{fig:boxplots} (b, d), Table \ref{tab:spearman_EC}), $N_C$ still showed a moderate upward trend ($\rho$ = 0.44, $p$ $<$ 0.05, Spearman rank correlation), suggesting that structural complexity continued to increase concomitant with fitness \citep{joshi2013minimal, edlund2011integrated, albantakis2014evolution}.

\begin{table*}[htb!]
\centering
\caption{Statistical comparisons (Mann–Whitney U test) for task performance and neural complexity between networks evolved with no energy costs (NEC) and with energy costs (EC) across environments with increasing seasonal dynamics (1 to 4 seasons). Comparisons are shown for overall task performance, network size ($N_S$), and structural complexity ($N_C$). '$==$' indicates no significant difference, and '$>$' indicates significantly greater than (given, $p$ $<$ 0.05).}
\begin{tabular}{|c|c|c|c|}
\hline
\multicolumn{1}{|l|}{} & \textbf{Task Performance} & \textbf{$\mathbf{N_S}$}                  & \textbf{$\mathbf{N_C}$}                  \\ \hline
\textbf{1 season}      & NEC == EC (p=0.27)        & NEC \textgreater EC (p=0.01) & NEC \textgreater EC (p=0.03) \\
\textbf{2 seasons}     & NEC == EC (p=0.54)        & NEC \textgreater EC (p=0.01) & NEC \textgreater EC (p=0.04) \\
\textbf{3 seasons}     & NEC == EC (p=0.30)        & NEC \textgreater EC (p=0.00) & NEC == EC (p=0.06)           \\
\textbf{4 seasons}     & NEC == EC (p=0.56)        & NEC \textgreater EC (p=0.00) & NEC \textgreater EC (p=0.00) \\ \hline
\end{tabular}
\label{tab:Energy_Impact}
\end{table*}

\begin{figure*}[t]
    \centering
    \includegraphics[width=0.545\linewidth]{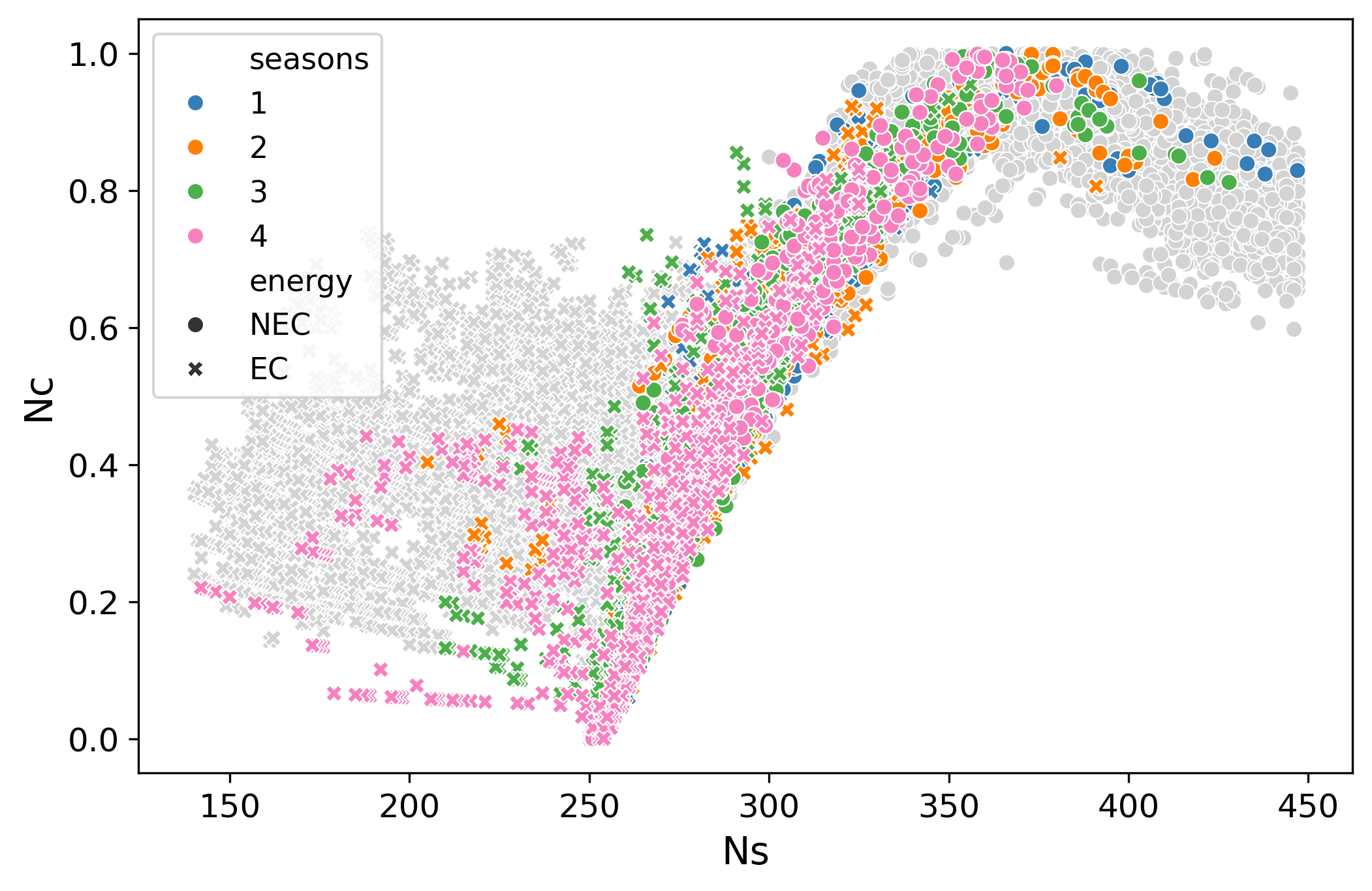}
    \caption{Network size ($N_S$) versus structural complexity ($N_C$) over generations. Grey points and crosses are random-walk evolution without and with an ANN size penalty, respectively. Colored markers show task-based evolution (1-4 season environments). Overlap across conditions suggests $N_C$ primarily reflects evolved $N_S$ given mutation rather than task driven selection.}
    \label{fig:scatterplot}
\end{figure*}

\subsection{Impact of Energy Costs on Neural Complexity and Task Performance}

Mann-Whitney U tests ($p$ $<$ 0.05) tested for significant differences in task performance and neural complexities ($N_S$ and $N_C$) between networks evolved with and without energy costs imposed on ANN size (NEC vs. EC scenarios), across each environment in the set. These tests (see Table \ref{tab:Energy_Impact}) revealed that, across all environments, imposing energy costs on ANN size led to the evolution of smaller ANNs with comparable task performance.

Structural complexity ($N_C$) was also significantly lower in the EC scenario compared to the NEC scenario, suggesting a potential dependency between $N_S$ and $N_C$. 
Overall, these results highlight the value of incorporating energy constraints to promote the evolution of more efficient neural architectures. Ongoing work is investigating the effects of varying energy costs to identify thresholds where energy limitations begin to significantly impact task performance.

\subsection{Evaluating the Dependency Between $\mathbf{N_S}$ and $\mathbf{N_C}$}

To examine whether structural complexity ($N_C$) was simply a byproduct of network size ($N_S$), 20 additional evolutionary runs were conducted using the same parameters as the previous experiments, but with fitness values assigned randomly from a uniform distribution $U(0,1)$, thereby removing task-based selection pressure. This approach evaluated whether randomly evolved ANNs, under the same evolutionary dynamics, exhibited similar $N_C$ values for a given $N_S$ (as those observed in the task-driven experiments).

Given $N_S$ decreased over generations in some task-driven evolutionary runs, a pattern not observed in the random-fitness runs, the additional random-walk evolutionary runs included an explicit penalty on ANN size to encourage smaller networks and enable a more direct comparison. The resulting scatter plot (Figure \ref{fig:scatterplot}) presents $N_S$ versus $N_C$ across all generations, with random-fitness runs in grey and task-driven runs in color. The colored points predominantly fall within the same bounds as the grey points, suggesting that $N_C$ may have emerged primarily as a consequence of evolving $N_S$ under the given mutation settings, rather than from task-driven pressures. Structural complexity, therefore, cannot be isolated as a key driver of performance in this context. This is likely a result of the limitations of the feedforward neural controllers \citep{Nolfi2000} used in this study's experiments. Ongoing work is testing recurrent connections (memory) in ANNs, to elucidate whether structural complexity plays a critical role in facilitating adaptive and energy efficient solutions across dynamic environments.

\section{Conclusions}
This study investigated how energy costs and changing environments influence the evolution of neural complexity in RL agent ANN controllers. Results indicated that changing environments only impacted neural complexity evolution when energy costs were imposed, with more seasonal environments driving the evolution of smaller networks. Results support the \textit{Expensive Brain Hypothesis} (EBH) over the \textit{Cognitive Buffer Hypothesis} (CBH), within the context of this foraging task, providing \textit{in silico} evidence that organisms in fluctuating environments may evolve smaller, more energy-efficient brains. Structural complexity increased with fitness, but is hypothesized to have emerged as a byproduct of evolution given current mutation settings. Moreover, imposing energy costs encouraged the evolution of more efficient ANNs, with implications for assisting the design of energy constrained robotic controllers \citep{nagar2019cost}, such as those that must adapt to changing robot morphologies and environments \citep{WatsonNitschke2015, MailerNitschkeRaw2021}. While these experimental environments allow controlled testing of energy costs and environmental variability, they are highly simplified, so generalization to neural evolution should be made with caution. Overall, this study's key contribution was its demonstration of the role of energy costs in shaping neural complexity.

\section{Acknowledgements}
Computations were performed using facilities provided by the University of Cape Town’s ICTS High Performance Computing team: \url{hpc.uct.ac.za}

\footnotesize
\balance
\bibliographystyle{apalike}
\bibliography{references}

\end{document}